\documentclass[letterpaper, 10 pt, conference]{ieeeconf}  

\IEEEoverridecommandlockouts                              

\usepackage{algorithm,algpseudocode}
\usepackage{amsmath, graphicx } 
\usepackage{bbold}
\usepackage{subcaption} 
\usepackage{breqn,bbm,xcolor}
\usepackage{multirow}
\usepackage[symbol]{footmisc}
\usepackage{hyperref}

\title{\LARGE \bf SuPer: A Surgical Perception Framework for Endoscopic Tissue Manipulation with Surgical Robotics}

\author{Yang Li$^{1*}$, Florian Richter$^{2*}$ \IEEEmembership{Student Member, IEEE} , Jingpei Lu$^2$, Emily K. Funk$^3$, \\ Ryan K. Orosco$^3$ \IEEEmembership{Member, IEEE}, Jianke Zhu$^1$ \IEEEmembership{Senior Member, IEEE} and Michael C. Yip$^{2}$ \IEEEmembership{Member, IEEE}
\thanks{$^{*}$Equal contributions. This work was done when Yang Li was a visiting student at the University of California San Diego.}%
\thanks{$^{1}$Yang Li and Jianke Zhu are with the College of Computer Science, Zhejiang University, Hangzhou, 310027 China. Jianke Zhu is also with the Alibaba-Zhejiang University Joint Research Institute of Frontier Technologies, Hangzhou, China.
{\tt\small\{liyang89, jkzhu\}@zju.edu.cn}}%
\thanks{$^{2}$Florian Richter, Jingpei Lu, and Michael C. Yip are with the Department of Electrical and Computer Engineering, University of California San Diego, La Jolla, CA 92093 USA.
{\tt\small\{frichter, jil360, yip\}@ucsd.edu}}%
\thanks{$^3$Emily K. Funk and Ryan K. Orosco is with the Department of Surgery - Division of Head and Neck Surgery, University of California San Diego, La Jolla, CA 92093 USA. 
{\tt\small \{ekfunk, rorosco\}@ucsd.edu }}%
}

\begin{document}

\maketitle
\thispagestyle{empty}
\pagestyle{empty}

\begin{abstract}
Traditional control and task automation have been successfully demonstrated in a variety of structured, controlled environments through the use of highly specialized modeled robotic systems in conjunction with multiple sensors.
However, the application of autonomy in endoscopic surgery is very challenging, particularly in soft tissue work, due to the lack of high-quality images and the unpredictable, constantly deforming environment.
In this work, we propose a novel surgical perception framework, SuPer, for surgical robotic control. This framework continuously collects 3D geometric information that allows for mapping a deformable surgical field while tracking rigid instruments within the field.
To achieve this, a model-based tracker is employed to localize the surgical tool with a kinematic prior in conjunction with a model-free tracker to reconstruct the deformable environment and provide an estimated point cloud as a mapping of the environment.
The proposed framework was implemented on the da Vinci Surgical\textregistered{} System in real-time with an end-effector controller where the target configurations are set and regulated through the framework.
Our proposed framework successfully completed soft tissue manipulation tasks with high accuracy.
The demonstration of this novel framework is promising for the future of surgical autonomy. 
In addition, we provide our dataset for further surgical research\footnote[2]{Website: \fontsize{6}{7}{\url{ https://www.sites.google.com/ucsd.edu/super-framework}}}.
\end{abstract}

\section{INTRODUCTION}

Surgical robotic systems, such as the da Vinci robotic platform\textregistered{}~(Intuitive Surgical, Sunnyvale, CA, USA), are becoming increasingly utilized in operating rooms around the world.
Use of the da Vinci robot has been shown to improve accuracy through reducing tremors and provides wristed instrumentation for precise manipulation of delicate tissue~\cite{reduce_tremors}.
Current innovative research has been conducted to develop new control algorithms for surgical task automation~\cite{yipDasJournal}. 
Surgical task automation could reduce surgeon fatigue and improve procedural consistency through the completion of tasks such as suturing~\cite{suturing}, cutting~\cite{surgical_cutting_rl}, and tissue debridement~\cite{debridement_removal}.

Significant advances have been made in surgical robotic control and task automation.
However, the integration of perception into these controllers is deficient even though the capabilities of surgical tool and tissue tracking technologies have advanced dramatically in the past decade.
Without properly integrating perception, control algorithms will never be successful in non-structured environments, such as those under surgical conditions.

\begin{figure}[t!]
\vspace{2mm}
\begin{subfigure}{0.24\textwidth}
\includegraphics[width=1\textwidth,height=1.1in]{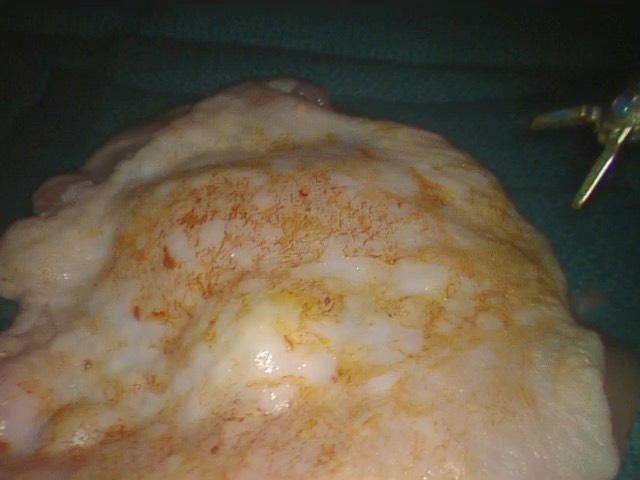}
\vspace{-0.12in}
\end{subfigure}
\begin{subfigure}{0.24\textwidth}
\includegraphics[width=1\textwidth,height=1.1in]{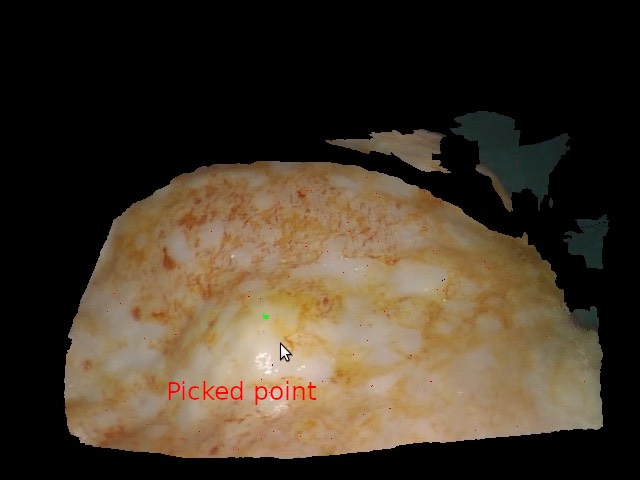}
\vspace{-0.12in}
\end{subfigure}
\begin{subfigure}{0.24\textwidth}
\includegraphics[width=1\textwidth]{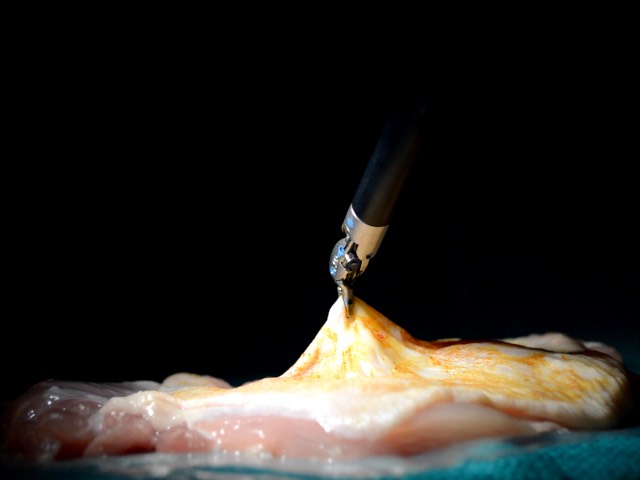}
\end{subfigure}
\begin{subfigure}{0.24\textwidth}
\includegraphics[width=1\textwidth]{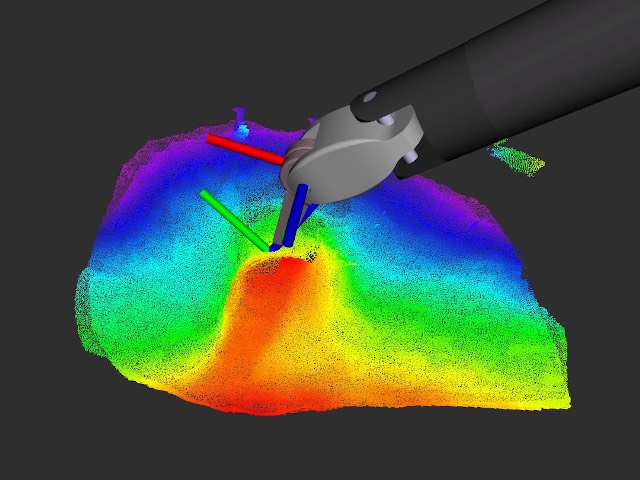}
\end{subfigure}

\caption{A demonstration of the proposed surgical perception framework. A green point on the perception model of the tissue, shown in top right, is selected by the user and the real surgical robot grasps and stretches the tissue at that location. As seen in the bottom two images, the framework is able to capture the tissue's deformation from the stretching.}
\label{fig:whole}

\vspace{-0.2in}
\end{figure}

In this work, we propose a novel Surgical Perception framework, SuPer, which integrates visual perception from endoscopic image data with a surgical robotic control loop to achieve tissue manipulation.
A vision-based tracking system is carefully designed to track both the surgical environment and robotic agents, e.g. tissue and surgical tool as shown in Fig.~\ref{fig:whole}.
However, endoscopic procedures have limited sensory information provided by endoscopic images and take place in a constantly deforming environment. 
Therefore, we separate the tracking system into two methodologies: model-based tracking to leverage the available kinematic prior of the agent and model-free tracking for the unstructured physical world.
With the proposed 3D visual perception framework, surgical robotic controllers can manipulate the environment in a closed loop fashion as the framework maps the environment, tracking the tissue deformation and localizing the agent continuously and simultaneously.
In the experimental section, we also demonstrate an efficient implementation of the proposed framework on a da Vinci Research Kit in which we successfully manipulate tissue.

To the best of our knowledge, the proposed perception framework is the first work to combine 3D visual perception algorithms for general control of a surgical robot in an unstructured, deforming environment.
More specifically, our contributions can be summarized as 
     1) a perception framework with both model-based tracking and model-free tracking components to track the tissue and localize the robot simultaneously,
     2) deformable environment tracking to track tissue from stereo-endoscopic image data,
     3) surgical tool tracking to accurately localize and control the surgical tool in the endoscopic camera frame, and
     4) a released data set of tissue manipulation with the da Vinci Surgical\textregistered{} System.
The framework is implemented on a da Vinci Surgical\textregistered{} System and multiple tissue manipulation experiments were conducted to highlight its accuracy and precision.
We believe that the proposed framework is a fundamental step toward endoscopic surgical autonomy in unstructured environments. 
With a uniform perception framework in the control loop, more advanced surgical task automation can be achieved.

\section{Related Works}
\label{sec:related work}
As the presented work is at the intersection of multiple communities, the related works are split into three sections.

\subsubsection{Deformable Reconstruction}
The first group of related works are from the 3D reconstruction or motion capture community~\cite{Ngo15iccv,Salzmann2010reconstructionbook,zhu2009fast,zhu2009nonrigid}.
Newcombe et al.~\cite{newcombe2011kinectfusion} proposed a real-time method for reconstruction of a static 3D model using a consumer-level depth camera based on volumes for their internal data structure, 
while Keller et al.~\cite{keller2013pointFusion} employed the use of surfel points rather than volumes.
The rigidness assumption was then removed to capture the motion of a deforming scene~\cite{newcombe2015dynamicfusion}.
To enhance the robustness of reconstruction, key-point alignment was added to the original cost function of the deformable reconstruction~\cite{innmann2016volumedeform}.
In addition, multiple-sensor approaches have shown to further improve accuracy~\cite{dou2016fusion4d}.
Guo et al.~\cite{gao18surfelwarp} achieved similar results for deformable object reconstruction with surfel points.

\subsubsection{Endoscopic Tissue Tracking}
Tissue tracking is a specific area of visual tracking that often utilizes 3D reconstruction techniques.
A comprehensive evaluation of different optical techniques for geometry estimation of tissue surfaces concluded that stereoscopic is the only feasible and practical approach to tissue reconstruction and tracking during surgery ~\cite{maier2014comparative}. 
For image-guided surgery, Yip et al.~\cite{yip2012tissueTracking} proposed a tissue tracking method with key-point feature detection and registration. 
3D dynamic reconstruction was introduced by Song et al.~\cite{song2017dynamicTissue} to track in-vivo deformations.
Meanwhile, dense SLAM methods~\cite{mahmoud2018live,marmol2019arthroSLAM} are applied to track and localize the endoscope in the surgical scene with image features.
In contrast with the algorithms mentioned above, our proposed framework not only tracks the surgical environment through deformable reconstruction, but also integrates the control loop of the surgical robotic arm for automation.
\subsubsection{Endoscopic Surgical Tool Tracking and Control} 

A recent literature survey by Bouget et al.~\cite{bouget2017vision} gave a detailed summary of image-based surgical tool detection. 
Markerless with tracking algorithms~\cite{kristan2015vot,li2019ldes,li2014samf} requires features which can be learned~\cite{original_rcs, Reiter}, generated via template matching~\cite{EKF}, or hand-crafted~\cite{Hao}.
After the features have been extracted, they are fused with kinematic information and encoder readings to fully localize the surgical robotic tools~\cite{data_fusion_surgical_tool_tracking}.

Once the surgical tool is localized, control algorithms can be applied on them to manipulate the environment.
Previous work in control algorithms for surgical robotics includes compliant object manipulation \cite{alambeigi2018robust}, debridement removal~\cite{dvrl, debridement_removal}, suture needle manipulation ~\cite{suturing, suture_needle_pick_up, zhong2019dual}, and cutting~\cite{surgical_cutting_rl, surgical_cutting}.
These control algorithms show advanced and sophisticated manipulations, however, they rely on structured environments and would have difficulties in the real surgical scene.

\section{METHODOLOGY}
\label{section:methodology}

\begin{figure}[t]
\vspace{2mm}
    \centering
    \includegraphics[width=0.9\linewidth]{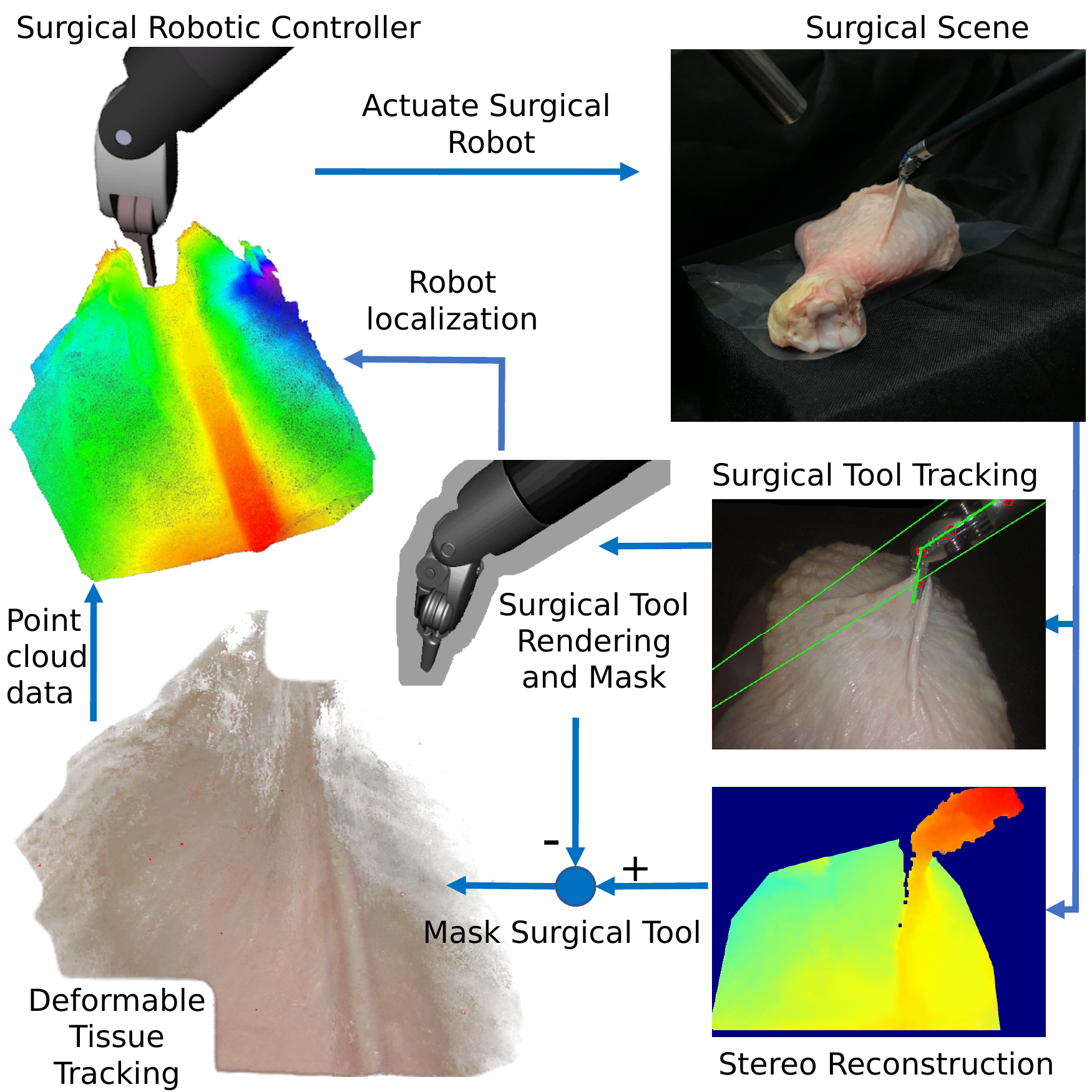}
    \caption{Flow chart of the proposed SuPer framework which integrates perception for localization and environment mapping into surgical robotic control.}
    \label{fig:flowchart}
    \vspace{-0.14in}
\end{figure}

The goal of the SuPer framework, as shown in Fig. \ref{fig:flowchart}, is to provide geometric information about the entire surgical scene including the robotic agent and the deforming environment.
A model-based tracker via particle filter is chosen to localize the surgical robotic tool by utilizing a kinematic prior and fusing the encoder readings and endoscopic image data.
For the surgical environment, a model-free deformable tracker is employed since the surgical environment is unstructured and constantly deforming. 
The model-free tracker uses the stereo-endoscopic data as an observation to reconstruct the deformable scene.
To efficiently combine the two separate trackers, a mask of the surgical tool is generated based on the surgical tool tracker and removed from the observation given to the model-free tracking component.
Since the trackers are both perceived in the same camera coordinate frame, a surgical robotic controller can be used in our SuPer framework to manipulate the unstructured surgical scene.

\subsection{Surgical Tool Tracking}

Surgical robots, such as the da Vinci\textregistered{} Surgical System, utilize setup-joints to position the base robotic arm and the endoscopic camera. These setup-joints have long links and therefore have large errors relative to the active joints during a procedure of the surgical robot~\cite{original_rcs, EKF}. Furthermore, calibration for the transform from the base of the robot to the camera, also known as the \textit{hand-eye} transform, rather than relying on the setup-joint kinematics, has been highlighted as unreliable when controlling surgical robots~\cite{cable_error_comp_nn}. Modeling this explicitly, a point on the $j$-th link, $\mathbf{p}^j \in \mathcal{R}^3$ is transformed to the camera frame:
\begin{equation}
    \label{eq:tool_tracking_formulation}
    \overline{\mathbf{p}}^c_t = \mathbf{T}^c_{b-} \mathbf{T}^{b-}_b \prod \limits_{i=1}^{j} \mathbf{T}_i^{i-1}(\theta^i_t) \overline{\mathbf{p}}^j
\end{equation}
where $\mathbf{T}^c_{b-}$ is the homogeneous hand-eye transform from calibration or the setup-joints, $\mathbf{T}^{b-}_b$ is the error in the hand-eye transform, and $\mathbf{T}_i^{i-1}(\theta^i_t)$ is the $i$-th homogeneous joint transform with joint angle $\theta^i_t$ at time $t$. Note that coordinate frame 0 is the base of the robot and that $\overline{\cdot}$ represents the homogeneous representation of a point (e.g. $\overline{\mathbf{p}} = [\mathbf{p}, 1]^T$). To track the surgical tools accurately, $\mathbf{T}^{b-}_b$ will be estimated in real-time. Similar problem formulations have been utilized in prior works for surgical tool tracking~\cite{original_rcs, Reiter, EKF}.

To track error, $\mathbf{T}_b^{b-}$ is parameterized by six scalar values: an axis-angle vector, $\mathbf{w} \in \mathcal{R}^3$, and a translational vector $\mathbf{b} \in \mathcal{R}^3$. The motion model, feature detection algorithm, and observation models are described in the remainder of this subsection. For implementation, we elected to use the particle filter because of its flexibility to model the posterior probability density with a finite number of samples~\cite{particle_filter_robotics}.

\subsubsection{Motion Model}
For initialization, the error of the hand-eye is assumed to be zero and the uncertainty of the calibration or setup-joints is modeled as Gaussian noise:
\begin{equation}
    [\mathbf{w}_{0|0}, \mathbf{b}_{0|0}]^T \sim \mathcal{N}([0, \dots, 0]^T, \mathbf{\Sigma}_{0} )
\end{equation}
where $ \mathbf{\Sigma}_{0}$ is the covariance matrix. Similarly, the motion model is set to have additive mean zero Gaussian noise since the particle filter is tracking the uncertainty in the hand-eye which is a constant transform:
\begin{equation}
    [\mathbf{w}_{t+1|t}, \mathbf{b}_{t+1|t}]^T \sim \mathcal{N}([\mathbf{w}_{t|t}, \mathbf{b}_{t|t}]^T, \mathbf{\Sigma_{w,b}} )
\end{equation}
where $\mathbf{\Sigma_{w,b}}$ is the covariance matrix.

\subsubsection{Features Detection and Camera Projections} 
As the focus of this work is not to develop surgical tool feature detection algorithms, we employ two simple feature approaches to verify our idea. Colored markers were drawn on the surgical tool to use as point features and the edges of the tool shaft are used as line features. The locations of the colored markers are similar to the detected features in Ye's et al. tool tracking work~\cite{EKF}. Please note that algorithms developed in previous literature can be utilized to robustly detect features from the endoscopic image data on the surgical tool to update the estimation for the particle filter~\cite{bouget2017vision}.

The painted markers were detected by converting the image to the Hue-Saturation-Value (HSV) color space and thresholding the hue channel. The mask generated from the thresholding is eroded and dilated to reduce the rate of small, false detections. Centroids, $\mathbf{m}^k_{t+1} \in \mathcal{R}^2$, are then calculated for each of the distinct contours of the mask to give a pixel point measurement of the markers. The camera projection equation for the detected pixel point of marker $i$ is:
\begin{equation}
    \hat{\mathbf{m}}_i(\mathbf{w}, \mathbf{b}) = \frac{1}{s} \mathbf{K} \mathbf{T}^c_{b-} \mathbf{T}^{b-}_b(\mathbf{w}, \mathbf{b}) \prod \limits_{i=1}^{j_i} \mathbf{T}_i^{i-1}(\theta^i_{t+1}) \overline{\mathbf{p}}^{j_i}_i
\end{equation}
where $\mathbf{p}^{j_i}_i \in \mathcal{R}^3$ is the known marker position on link $j_i$ and $\frac{1}{s}\mathbf{K}$ is the standard camera projection operation and $\mathbf{K}$ is the intrinsic camera calibration matrix.

The second feature detected is the projected edges of the insertion shaft of the surgical tool, which is a cylinder. Pixels potentially associated with the edges are detected using Canny edge detector~\cite{canny_edge_detector} and classified into distinct lines using the Hough transform~\cite{hough_transform}. This results in a list of detected lines parameterized by scalars $\rho^k_{t+1}$ and $\phi^k_{t+1}$:
\begin{equation}
    \rho^k_{t+1} = u \textrm{ cos} (\phi^k_{t+1}) + v \textrm{ sin}(\phi^k_{t+1}) \label{equ:line}
\end{equation}
where $u$ and $v$ are pixel coordinates. For the sake of brevity, the camera projection equations for a cylinder resulting in two lines is omitted. Please refer to Chaumette's work for a full derivation and expression~\cite{project_cylinder}. The camera projection equation for a single line $i$ is denoted as $\hat{\rho}_i(\mathbf{w}, \mathbf{b})$ and $\hat{\phi}_i(\mathbf{w}, \mathbf{b})$ using the same parameterization as (\ref{equ:line}).

\subsubsection{Observation Model}
To make associations between the detected marker features, $\mathbf{m}^k_t$, and their corresponding marker, a greedy matching technique is done because of the low computation time. An ordered list of the cost
\begin{equation}
    C^m_{k,i}(\mathbf{w},\mathbf{b}) = e^{-\gamma_m || \mathbf{m}^k_t - \hat{\mathbf{m}}_i(\mathbf{w}, \mathbf{b}) ||^2}
\end{equation}
for detection $k$ and projected marker $i$ is made where $\gamma_m$ is a tuned parameter. Iteratively, detection $k$ and marker $i$ from the lowest value of this cost list is matched, the tuple is added to the associated data list $A_m$, and all subsequent costs associated with either $k$ or $i$ are removed from the list. This is done until a max cost, $C^m_{max}$, is reached.

The same procedure is utilized for the detected lines $[\rho^k_t, \phi^k_t]$,  and the projected edges of the insertion shaft except the cost equation is
\begin{equation}
        C^l_{k,i}(\mathbf{w},\mathbf{b}) = e^{-\gamma_\phi |\phi^k_t -  \hat{\phi}_i(\mathbf{w}, \mathbf{b})| - \gamma_\rho | \rho^k_t -  \hat{\rho}_i(\mathbf{w}, \mathbf{b}) |}
\end{equation}
where $\gamma_\phi$ and $\gamma_\rho$ are tuned parameters, the data list is denoted as $A_l$, and a max cost of $C^l_{max}$.

\begin{figure}[t]
    \centering
    \vspace{2mm}
    \begin{subfigure}{0.15\textwidth}
        \includegraphics[width=\linewidth]{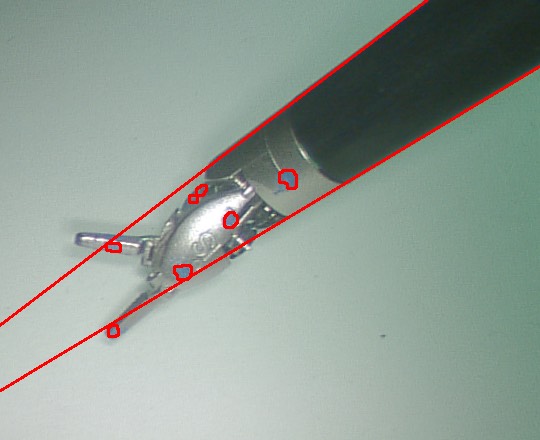}
    \end{subfigure}%
    \hspace{0.3mm}
    \begin{subfigure}{0.15\textwidth}
        \includegraphics[width=\linewidth]{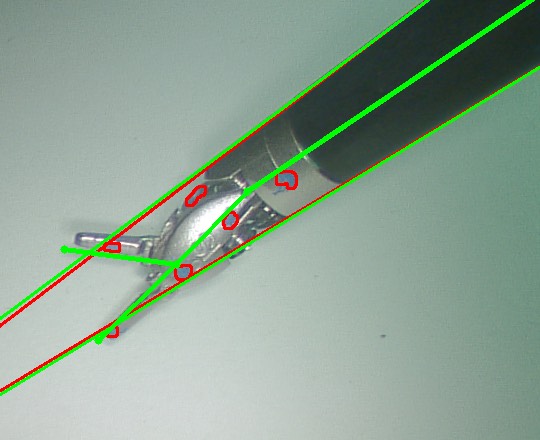}
    \end{subfigure}%
    \hspace{0.3mm}
    \begin{subfigure}{0.15\textwidth}
        \includegraphics[width=\linewidth]{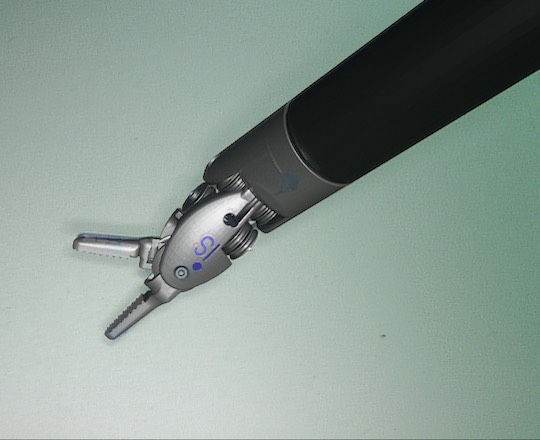}
    \end{subfigure}
    
    \vspace{1mm}
    
    \caption{Surgical tool tracking implementation on the da Vinci Surgical\textregistered{} System running 30fps in real-time. From left to right the figures show: detected markers and edges, re-projected kinematic tool and shaft edges, and the full Augmented Reality rendering of the surgical tool~\cite{render_psm} on top of the raw endoscopic data (best viewed in color).}
    \vspace{-0.14in}
    \label{fig:tool_tracking}
\end{figure}

The association costs are wrapped in a radial basis function so they can be directly used for the observation models. The probability of the detected markers, $\mathbf{m}_{t+1}$, is modeled as:
\begin{multline}
    P( \mathbf{m}_{t+1}  | \mathbf{w}_{t+1|t}, \mathbf{b}_{t+1|t}) \propto (n_m - |A_m|)C^m_{max} \\ + \sum \limits_{ k,i  \in A_m } C^m_{k,i} (\mathbf{w}_{t+1|t}, \mathbf{b}_{t+1|t})
\end{multline}
where there are a total of $n_m$ markers painted on the surgical tool. Similarly, the probability of the detected lines, $\boldsymbol{\rho}_{t+1}$, $\boldsymbol{\phi}_{t+1}$, is modelled as
\begin{multline}
    P( \boldsymbol{\rho}_{t+1}, \boldsymbol{\phi}_{t+1} |\mathbf{w}_{t+1|t}, \mathbf{b}_{t+1|t}) \propto (2 - |A_l|)C^l_{max}  \\ + \sum \limits_{ k,i \in A_m } C^l_{k,i} (\mathbf{w}_{t+1|t}, \mathbf{b}_{t+1|t})
\end{multline}
where the max $|A_l|$ is two since there is only one cylinder from the instruments shaft. These functions are chosen since they increase the weight of a particle for stronger associations, but does not completely zero out the weight if no associations are made which can occur in cases of obstruction or missed detections. Since these two observations occur synchronously, the update is combined using the switching observation models-synchronous case~\cite{multiple_obs_pf}. Example images of the tool tracking are shown in Fig.~\ref{fig:tool_tracking}.

\subsection{Depth Map from Stereo Images}
The depth map from the stereoscopic image is generated using the Library for Efficient Large-Scale Stereo Matching (LIBELAS)~\cite{stereo_matching}. 
To fully exploit the prior and enhance the robustness of our system,
the surgical tool portion of the image and depth data is not passed to the deformable tissue tracker since the surgical tool is already being tracked.
Therefore, a mask of the surgical tool is generated using the same OpenGL rendering pipeline we previously developed~\cite{render_psm}, and applied to the depth and image data passed to the deformable tissue tracker. To ensure the mask covers all of the tool, it is dilated before being applied.

\subsection{Deformable Tissue Tracking}

To represent the environment, we choose surfel~\cite{keller2013pointFusion} as our data structure due to the direct conversion to point cloud which is a standard data type for the robotics community.
A surfel $\mathcal{S}$ represents a region of an observed surface and is parameterized by the tuple $(\mathbf{p}, \mathbf{n}, \mathbf{c}, \mathbb{r}, \mathbb{c}, \mathbb{t})$, where $\mathbf{p}, \mathbf{n}, \mathbf{c} \in \mathcal{R}^3$ are the expected position, normal, and color respectively and scalars $\mathbb{r}, \mathbb{c}, \mathbb{t}$ are the radius, confidence score, and time stamp of last update respectively. 
Alongside the geometric structure the surfel data provides, it also gives confidence and timestamp of the last update which can be exploited to further optimize a controller working in the tracked environment.
For adding/deleting and fusing of surfels, refer to work done by Keller et al.~\cite{keller2013pointFusion} and Gao et al.~\cite{gao18surfelwarp}.



\subsubsection{Driven/Parameterized Model}
\begin{figure}[t!]
\vspace{2mm}
\begin{subfigure}{0.12\textwidth}
\includegraphics[width=1\textwidth]{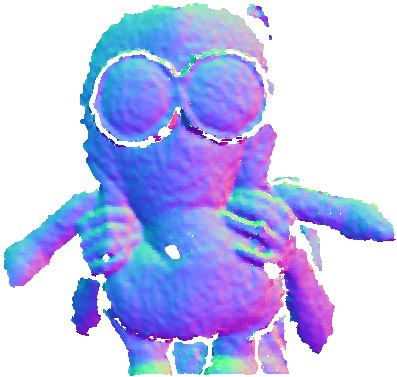}
\label{fig:subim1}
\end{subfigure}
\begin{subfigure}{0.12\textwidth}
\includegraphics[width=0.95\textwidth]{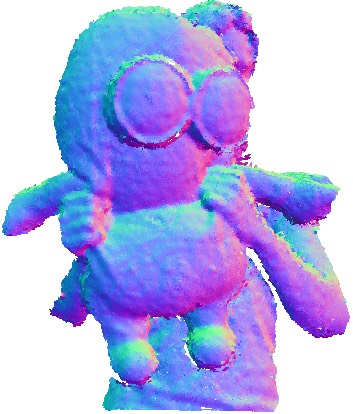}
\label{fig:subim2}
\end{subfigure}
\begin{subfigure}{0.115\textwidth}
\includegraphics[width=1\textwidth]{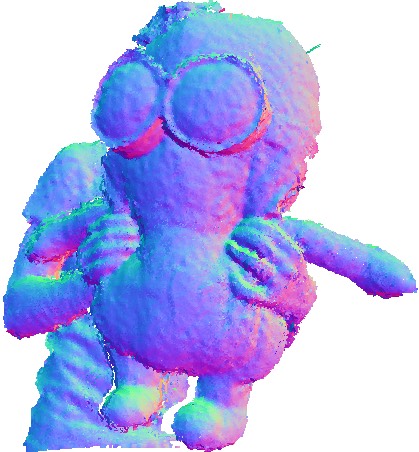}
\label{fig:subim3}
\end{subfigure}
\begin{subfigure}{0.115\textwidth}
\includegraphics[width=0.95\textwidth]{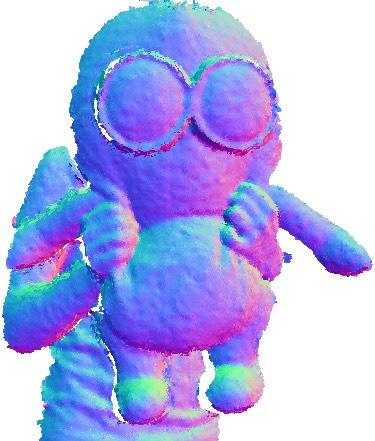}
\label{fig:subim4}
\end{subfigure}
\caption{Deformable tracking results with testing dataset~\cite{innmann2016volumedeform}. The color represents the normal of our surfel data. As the model fuses with more data from left to right, the normal becomes smooth and the deformations are captured.}
\vspace{-0.14in}
\label{fig:minion_deformable_tracker}
\end{figure}

The number of surfels grows proportionally to the number of image pixels provided to the deformable tracker, so it is infeasible to track the entire surfel set individually.
Inspired by the work of Embedded Deform~(ED)~\cite{sumner2007embedded}, we drive our surfel set with a less-dense ED graph, $\mathcal{G}_{ED} = \{\mathcal{V},\mathcal{E},\mathcal{P}\}$, where $\mathcal{V}$ is the vertex index set, $\mathcal{E}$ is the edge set and $\mathcal{P}$ is the parameters set. With a uniform sampling from the surfel, the number of ED nodes, $N_{ED}$, is much fewer than the number of surfels, $N_{surfel}$. 
Thus, the ED graph has significantly fewer parameters to track compared with the entire surfel model. 
Moreover, the ED graph can be thought of as an embedded sub-graph and skeletonization of the surfels to capture their deformations.
The transformation of every surfel is modeled as follows:
\begin{equation}
\label{transforPos}
    T(\mathbf{\overline{p}}) = \mathbf{T}_{g}\sum_{i \in \text{KNN}(\mathbf{\overline{p}})} \alpha_i[\mathbf{T}(\mathbf{q}_i,\mathbf{b}_i)(\overline{\mathbf{p}} - \vec{ \mathbf{g}_i}) + \vec{\mathbf{g}_i} ]
\end{equation}
where $\mathbf{T}_g$ is the global homogeneous transformation (e.g. common motion shared with all surfel), $\alpha_i$ is a normalized weight, and KNN$(\mathbf{\overline{p}})$ is an index set that contains $k$-nearest neighbors of $\mathbf{\overline{p}}$ in $\mathcal{G}_{ED}$.
An ED node consists of a parameter tuple $(\mathbf{g}_i,\mathbf{q}_i,\mathbf{b}_i) \in \mathcal{P}$ where $\mathbf{g}_i \in \mathcal{R}^3$ is the position of the ED node and $\mathbf{q}_i \in \mathcal{R}^4$ and $\mathbf{b}_i \in \mathcal{R}^3$ are the quaternion and translation parameters respectively and converted to a homogeneous transform matrix with $\mathbf{T}(\mathbf{q}_i,\mathbf{b}_i)$. Both $\alpha_i$ and KNN$(\mathbf{\overline{p}})$ are generated using the same method proposed by Sumner et al.~\cite{sumner2007embedded}. Note that $\vec{\cdot}$ is a vector in homogeneous representation(e.g. $\vec{\mathbf{g}} = [\mathbf{g}, 0]^T$). The normal transformation is similarly defined as:
\begin{equation}
\label{transforNorm}
    T_n(\mathbf{n}) = \mathbf{T}_{g}\sum_{i \in \text{KNN}(\mathbf{\overline{p}})} \alpha_i[\mathbf{T}(\mathbf{q}_i, \mathbf{0})\vec{\mathbf{n}}]
\end{equation}
When implementing the ED graph, the $\mathbf{q}_i$ and $\mathbf{b}_i$ for node $i$ are the current frames estimated deformation. After every frame, the deformations are committed to $\mathbf{g}_i$ and the surfels based on (\ref{transforPos}) and (\ref{transforNorm}). Therefore, with an ED graph of $n$ nodes, the whole surfel model is estimated with $7 \times (n + 1)$ parameters. Note that the extra 7 parameters come from $\mathbf{T}_{g}$ which is also estimated with a quaternion and translational vector. An example of using this model to track deformations is shown in Fig. \ref{fig:minion_deformable_tracker}.

\subsubsection{Cost Function}
To track the visual scene with the parameterized surfel model, a cost function is defined to represent the distance between an observation and the estimated model. It is defined as follows:
\begin{equation}
\label{eq:costfunction}
    E = E_{data} + \lambda_a E_{Arap} + \lambda_r E_{Rot} + \lambda_c E_{Corr}
\end{equation}
where $E_{data}$ is the error between the depth observation and estimated model, $E_{ARAP}$ is a rigidness cost such that ED nodes nearby one another have similar deformation, $E_{Rot}$ is a normalization term for the quaternions to satisfy a rotation in $SO_3$ space, and $E_{Corr}$ is a visual feature correspondence cost to ensure texture consistency.

More specifically, the traditional point-plane error metric~\cite{newcombe2011kinectfusion} is used for the depth data cost. When minimized, the model is aligned with the observed depth image. The expression is:
\begin{equation}
    E_{data} = \sum_i(T_n(\vec{\mathbf{n}}_{i})^T( T(\overline{\mathbf{p}}_{i}) - \overline{\mathbf{o}}_i))^2
\end{equation}
where $\mathbf{o}_i = D(u,v)\mathbf{K}^{-1}[u,v,1]^T$ is the observed position from the depth map, $D$ at pixel coordinate $(u,v)$, and $\mathbf{p}_{i}$ and $\mathbf{n}_{i}$ are the associated surfel position and normal from the most up to date model. 
This cost term, however, is highly curved and not easy to solve.
To simplify the optimization, the normal is fixed at every iteration during optimization.
This results in the following expression at iteration $j$: 
\begin{equation}
    E^{(j)}_{data} = \sum_{i}(\hat{\mathbf{n}}^{(i)T}_{i}( T(\overline{\mathbf{p}}_{i};O^{(j)}) - \overline{\mathbf{o}}_i))^2
\end{equation}
where $\hat{\mathbf{n}}^{(j)}_{i} = T_n(\vec{\mathbf{n}}_{i-1};O^{(j-1)})$ and $O^{(j)}$ is the set of ED nodes at iteration $j$. This is a normal-difference cost term similar to Iterative Closest Point~\cite{newcombe2011kinectfusion}.


The rigid term is constructed by $l_2$ norm of the difference between the positions of an ED node transformed by two nearby transformations. The cost expression is:
\begin{equation}
   E_{ARAP} = \sum_i \sum_{k \in \mathbf{e}_i}||T(\mathbf{q}_k,\mathbf{b}_k)(\overline{\mathbf{g}}_i - \vec{\mathbf{g}}_k) + \vec{\mathbf{g}}_k  - \vec{\mathbf{g}}_i -\vec{\mathbf{b}}_i ||^2
\end{equation}
where $\mathbf{e}_i \in \mathcal{E}$ is the edge set of ED nodes neighboring node $i$. The edge set $\mathcal{E}$ is generated by the k-nearest neighbor algorithm based on ED node positions.
This cost term forces the model to have consistent motion among the nearby ED nodes. Intuitively, it gives hints to the model when a portion of the ED nodes do not receive enough data from the observation in the current frame.

To have a rigid-like transformation, the normalizing term in the cost function is set to:
\begin{equation}
    E_{Rot} = \sum_{k}||1 - \mathbf{q}_k^T\mathbf{q}_k||^2
\end{equation}
since quaternions hold $ ||\mathbf{q}||^2 = 1$.
Both $E_{Rot}$ and $E_{ASAP}$ are critical to ensuring all ED nodes move as rigid as possible. 
This is since $7 \times n$ is a very large space to optimize over relative to the observed data. For example, in cases of obstruction, the optimization problem is ill-defined without these terms.


The final cost term is for visual feature correspondence to force visual texture consistency between the model and the observed data. The expression for the cost is:
\begin{equation}
    E_{Corr} = \sum_{(\mathbf{m},\mathbf{c}) \in Feat}||T(\overline{\mathbf{p}}_{\mathbf{m}}) - \overline{\mathbf{o}}_{\mathbf{c}}||^2
\end{equation}

where $Feat$ is a set of associated pairs of matched feature points $\mathbf{m}, \mathbf{c} \in \mathcal{R}^2$ between the rendered color image of our model and the observed color image data respectively.
The observed point is obtained using the same expression as before: $\mathbf{o}_c = D(\mathbf{c})\mathbf{K}^{-1}\overline{\mathbf{c}}$. The feature matching gives a sparse but strong hint for the model to fit the current data.

\subsubsection{Optimization solver}
To solve the non-linear least square problem proposed in (\ref{eq:costfunction}), the Levenberg Marquardt~(LM) algorithm~\cite{book1992LMsolver} is implemented to efficiently obtain the solution for the model. 
The LM algorithm requires the cost function to be in the form of a sum of squared residuals.
Therefore, all the parameters from $O$ are stacked into a vector, $\mathbf{x}$, and all cost terms are reorganized into vector form such that $ ||\mathbf{f}(\mathbf{x})||^2= \mathbf{f}(\mathbf{x})^T\mathbf{f}(\mathbf{x})=E$. 
In this form, the function is linearized with a Taylor expansion:
\begin{equation}
    \delta =  \arg\min_{\substack{\delta}} ||\mathbf{f}(\mathbf{x}) + \mathbf{J}\delta||^2 
\end{equation}
where $\mathbf{J}$ is the Jacobian matrix of $\mathbf{f}(\mathbf{x})$. 
Following the LM algorithm, $\delta$ is solved for by using:
\begin{equation}
    \label{eq:pcg}
    (\mathbf{J}^T\mathbf{J}+\mu\mathbf{I})\delta = \mathbf{J}^T\mathbf{f}(\mathbf{x})
\end{equation}
where $\mu$ is a damping factor.
The LM algorithm accepts the $\delta$ by setting $\mathbf{x} \leftarrow \mathbf{x} + \delta$ when the cost function decreases: $||\mathbf{f}(\mathbf{x})||^2>||\mathbf{f}(\mathbf{x}+\delta)||^2$.
Otherwise, it increases the damping factor.
Intuitively, the LM algorithm tries to find a balance between the Gaussian-Newton method and the gradient descent solver. 
In our implementation, (\ref{eq:pcg}) is solved with a GPU version of the preconditioned conjugate gradient method within 10 iterations. 


\section{Experiments}
\label{section:experiment}
To measure the effectiveness of the proposed framework, our implementation was deployed on a da Vinci Surgical\textregistered{} System. The stereo camera is the standard 1080p laparoscopic camera running at 30fps. The Open Source da Vinci Research Kit (dVRK)~\cite{dvrk} was used to send end-effector commands and get joint angles and the end-effector location in the base frame of a single surgical robotic arm with a gripper, also known as Patient Side Manipulator (PSM). The data for the PSM is being sent at a rate of 100Hz. All of the communication between subsystems of the code was done using the Robot Operating System~(ROS), and everything ran on two identical computers with an Intel\textregistered{} Core\texttrademark{} i9-7940X Processor and NVIDIA's GeForce RTX 2080.

\subsection{Implementation Details}
Details for implementation of the proposed framework on the dVRK are stated below and organized by the components of the framework.

\subsubsection{Surgical Tool Tracking}
The particle filter used $N = 500$ particles, bootstrap approximation for the prediction step, and stratified resampling when the number of effective particles dropped below $N_{eff} = 200$ to avoid particle depletion. 
For initialization, the covariance, $\mathbf{\Sigma}_0$ is set to diag($0.025, 0.025, 0.025, 0.1, 0.1, 0.1$) where $\mathbf{w}$ is in radians and $\mathbf{b}$ is in mm. 
The motion model covariance, $\mathbf{\Sigma_{w,b}}$, is set to $0.1(\mathbf{\Sigma}_0)$. For the observation model, $[\gamma_m, \gamma_{\phi},\gamma_{\rho}] = [0.01, 10.0, 0.05]$ and $[C^m_{max}, C^l_{max}] = [e^{-50\gamma_m}, e^{-0.15\gamma_\phi  - 75\gamma_\rho} ]$. 
The image data is resized to 960 by 540 before processing for features. 
For the initial hand-eye transform, $\mathbf{T}_{b-}^c$, OpenCV's perspective-n-point solver is used on the segmented centroids of the markers.

\subsubsection{Depth Map from Stereo Images}
The endoscopic image data is resized to 640 by 480 before processing. 
The LIBELAS parameters used are the default settings from its open-sourced repository~\cite{stereo_matching}. 
After computing the depth map, $D$, it is masked by the rendered surgical tool. 
The mask is dilated by 9 pixels before being applied.
The depth map is then smoothed spatially with a bilateral filter and temporally with a median filter of four frames to decrease noise.


\subsubsection{Deformable Tracking}
The surfel radius is set to $\mathbb{r} = \sqrt{2}D(u,v)/(f|n_z|)$ and confidence score is calculated with $\mathbb{c} = exp(-d_c^2/0.72)$ at pixel coordinate $(u,v)$ where $n_z$ is the z component of camera frame normal, $f$ is the cameras focal length, and $d_c$ is the normalized distance from the pixel coordinate to the center of the image~\cite{newcombe2011kinectfusion}\cite{keller2013pointFusion}. 
Whenever new surfels are added to the model, ED nodes are randomly sampled from them~\cite{gao18surfelwarp}.
This typically results in 300 ED nodes, and therefore roughly 2K parameters to estimate.
Very similar surfels, both temporally and spatially, are merged to each other when we fuse the observed map to the model to keep the model concise as described in~\cite{gao18surfelwarp}.
OpenCV's implementation of SURF is used for feature extraction and matching in the cost functions visual correspondence term.
For the cost function, the parameters $[\lambda_{a}, \lambda_{r}, \lambda_{c}]$ are set to $[10, 100, 10 ]$.





\def \trimFactor {0.2}

\begin{figure*}[t!]
\centering
\vspace{2mm}
\begin{subfigure}{\trimFactor\textwidth}
\includegraphics[width=1\textwidth]{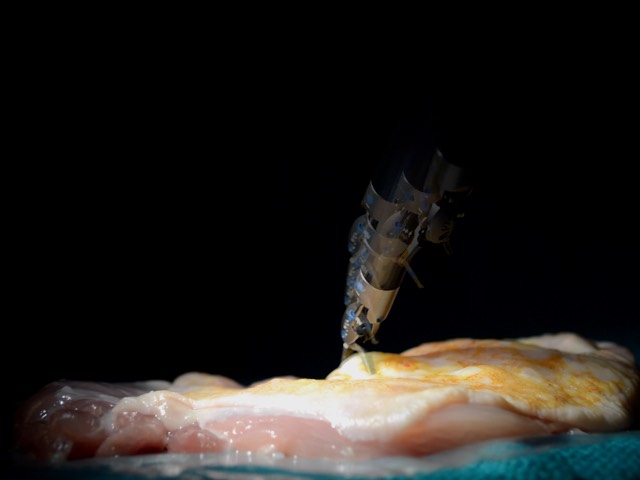}
\vspace{-0.12in}
\end{subfigure}
\begin{subfigure}{\trimFactor\textwidth}
\includegraphics[width=1\textwidth]{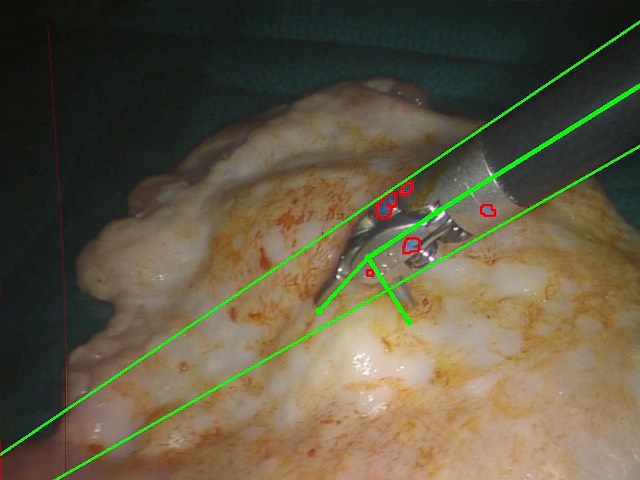}
\vspace{-0.12in}
\end{subfigure}
\begin{subfigure}{\trimFactor\textwidth}
\includegraphics[width=1\textwidth]{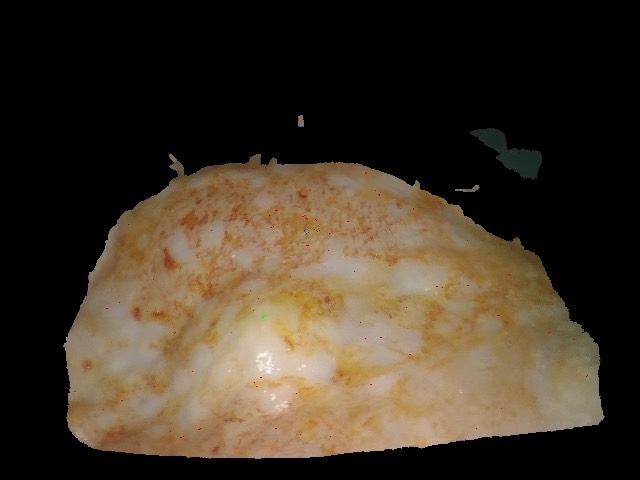}
\vspace{-0.12in}
\end{subfigure}
\begin{subfigure}{\trimFactor\textwidth}
\includegraphics[width=1\textwidth]{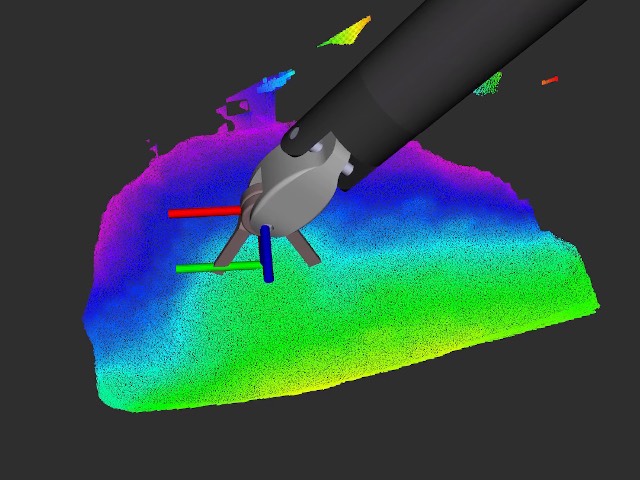}
\vspace{-0.12in}
\end{subfigure}

\begin{subfigure}{\trimFactor\textwidth}
\includegraphics[width=1\textwidth]{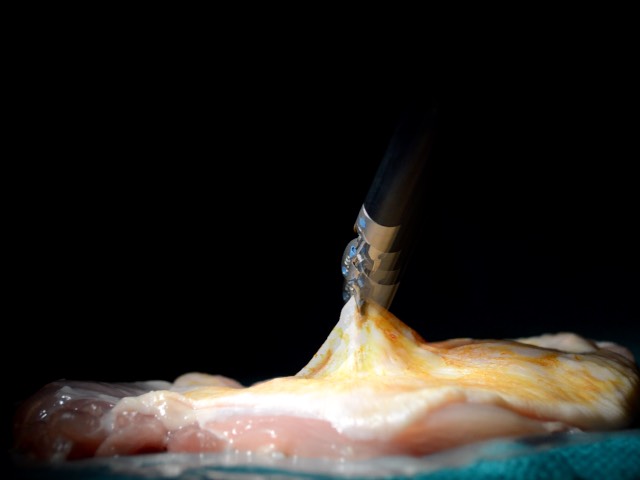}
\vspace{-0.12in}
\end{subfigure}
\begin{subfigure}{\trimFactor\textwidth}
\includegraphics[width=1\textwidth]{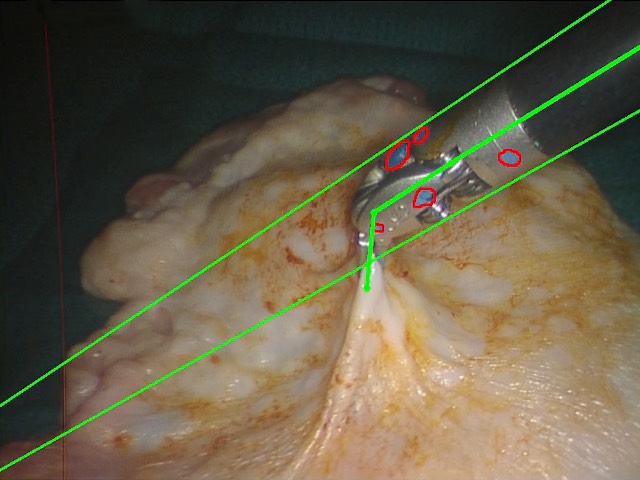}
\vspace{-0.12in}
\end{subfigure}
\begin{subfigure}{\trimFactor\textwidth}
\includegraphics[width=1\textwidth]{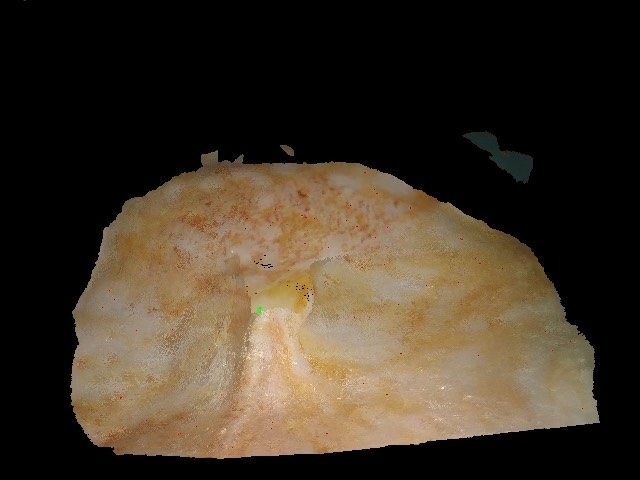}
\vspace{-0.12in}
\end{subfigure}
\begin{subfigure}{\trimFactor\textwidth}
\includegraphics[width=1\textwidth]{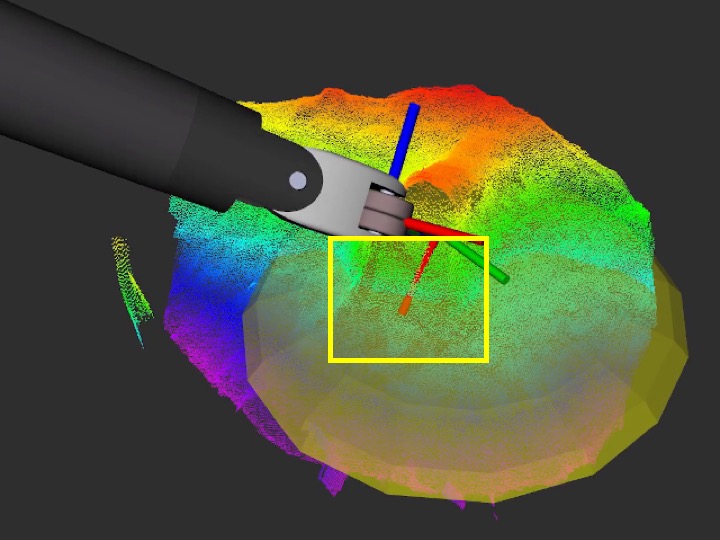}
\vspace{-0.12in}
\end{subfigure}

\caption{Tissue manipulating with the proposed SuPer framework implemented on the da Vinci\textregistered{} Surgical System in real-time. From left to right the figures show: the real scene, tool tracking from the endoscopic camera, deformable reconstruction, and RViz with point cloud of the environment, robot localization, and the tracked point to grasp. }
\label{fig:full_system_results}
\end{figure*}

\def \trimFactorT {0.22}
\begin{figure*}[t!]
\center
\begin{subfigure}{0.214\textwidth}
\includegraphics[width=1\textwidth]{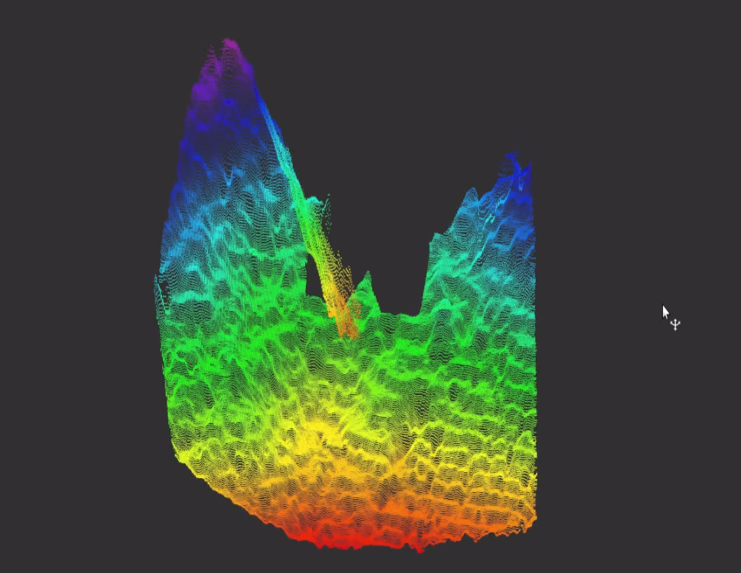}
\caption{Raw depth map}
\label{fig:depthMap}
\end{subfigure}
\begin{subfigure}{\trimFactorT\textwidth}
\includegraphics[width=1\textwidth]{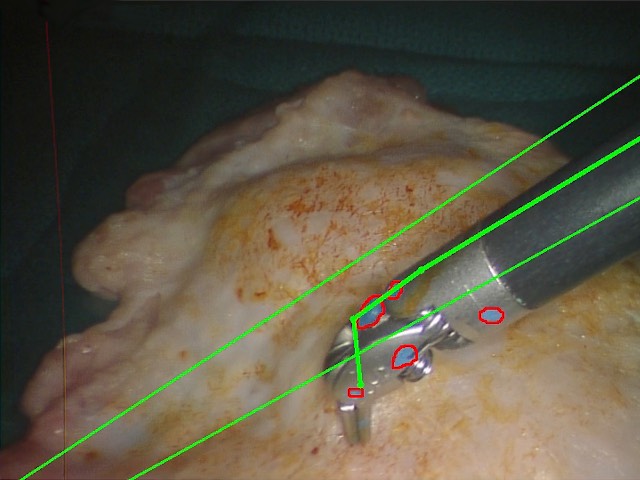}
\caption{Without tool tracking}
\end{subfigure}
\begin{subfigure}{\trimFactorT\textwidth}
\includegraphics[width=1\textwidth]{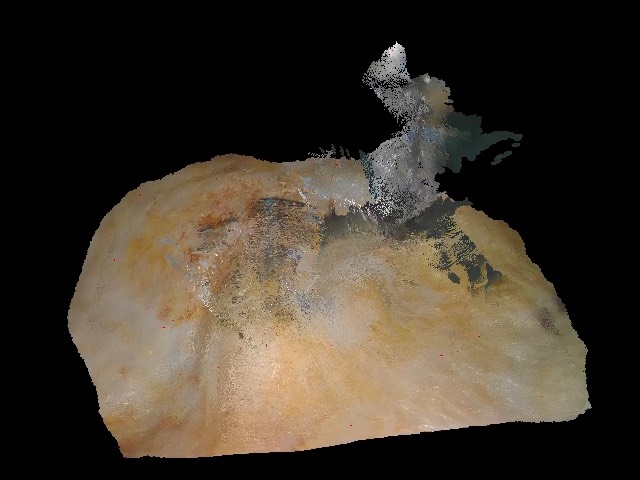}
\caption{Without mask}
\end{subfigure}
\begin{subfigure}{\trimFactorT\textwidth}
\includegraphics[width=1\textwidth]{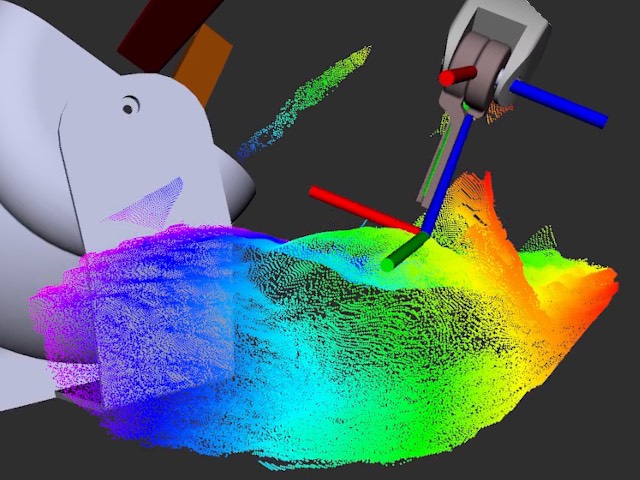}
\caption{w/o deformable tracking}
\end{subfigure}
\caption{Results from the repeated tissue manipulation experiment without using the complete proposed SuPer framework. None of these results are ideal since they do not properly capture the real surgical scene through failed robotic localization or improper environmental mapping.}
\label{fig:failure_results}
\vspace{-5mm}
\end{figure*}

\subsection{Repeated Tissue Manipulation}

To test the effectiveness of the proposed framework, a simple controller was implemented to grasp and tug on the tissue at the same tracked point repeatedly. 
At the beginning of the experiment, a small cluster of surfels is selected on the tissue in the deformable tracker, and their resulting averaged position, $\mathbf{p}^c_g$, and normal, $\mathbf{n}^c_g$, is the tracked point to be grasped. 
The following steps are then repeated five times or until failure on the PSM gripper.
    1) Align above surface: move to $\mathbf{p}^c_g + d\mathbf{n}^c_g$ where $d=2$cm and orientation $\mathbf{q}^c_g$ such that the opening of the gripper is pointed towards the surface normal, $\mathbf{n}^c_g$
    2) Move to the tissue: stop updating $\mathbf{p}^c_g$ and $\mathbf{n}^c_g$ from the deformable tracker and move to $\mathbf{p}^c_g + d\mathbf{n}^c_g$ where $d=0.5$cm and orientation $\mathbf{q}^c_g$
    3) Grasp and stretch the tissue: close the gripper to grasp the tissue and move to $\mathbf{p}^c_g + d\mathbf{n}^c_g$ where $d=2$cm and orientation $\mathbf{q}^c_g$
    4) Place back the tissue: move to $\mathbf{p}^c_g + d\mathbf{n}^c_g$ where $d=0.5$cm and orientation $\mathbf{q}^c_g$ and open the gripper
    5) Continue updating $\mathbf{p}^c_g$ and $\mathbf{n}^c_g$ from the deformable tracker.
Note that the end-effector on the PSM gripper is defined on the link preceding the jaws from the gripper which are approximately 1cm long.

To move the PSM to the target end-effector position, $\mathbf{p}^c_g + d\mathbf{n}^c_g$, and orientation, $\mathbf{q}^c_g$, trajectories are generated using linear and spherical linear interpolation respectively. 
The trajectories are re-generated after every update to $\mathbf{p}^c_g$ and $\mathbf{n}^c_g$ from the deformable tracker and generated in the camera frame from the current end-effector pose. 
The current end-effector pose is calculated by transforming the PSM end-effector pose from dVRK with the hand-eye transform from the surgical tool tracker. 
Finally, to follow the trajectory, the end-effector poses are transformed back to the base frame of the PSM using the surgical tool tracker and set via dVRK.

This experiment is repeated with these configurations:
    1) The complete proposed framework.
    2) The framework without deformable tracking, just static reconstruction, by setting the number of ED nodes to 0.
    3) The framework without surgical tool masking.
    4) The framework without surgical tool tracking, and instead relying on calibrated hand-eye.
The tissue used is the skin of a chicken leg.

\subsection{Reprojection Error for Tracking Accuracy}

To evaluate our proposed approach quantitatively, we manually annotated 20 points on the tissue through time on the raw image data from the repeated tissue manipulation experimentation. 
The 20 points are chosen from the highest confidence points of SURF in the first frame. 
This time series of 2D image positions is compared against the reprojection from the deformable tissue tracker. We also evaluate the result of an off-the-shelf SURF approach from OpenCV which matches the key points in every frame with the description in the first frame.
Moreover, the surgical tool accuracy is evaluated by comparing 50 manually segmented images, selected at random, from the repeated tissue experiment and compared against the reprojected/rendering of the surgical tool tracking. 
The experiment was conducted with different numbers of particles to highlight the trade-off between the accuracy of modeling the posterior probability and computational cost in real-time tracking.



\section{Results}
\label{section:results}




The separate components of the framework ran at 30fps, 30fps, 8fps, and 3fps for the surgical tool tracking, surgical tool rendering, depth map generation, and deformable tissue tracker respectively. 
An example of the procedure used for the repeated tissue manipulation experiment is shown in Fig. \ref{fig:whole}. 
When using the complete framework, the PSM arm successfully grasped the same location of the tissue all five times after repeated deformations.
As shown by the yellow rectangle in Fig. \ref{fig:full_system_results},
the deformable tracker even managed to capture the structure of the tissue that was not visible to the endoscopic camera during stretching.

When not using the deformable tracker, the computer crashed due to memory overflow after three grasps and the reconstruction was not at all representative of the real environment. 
With no mask, the reconstructed scene in the deformable tracker was unable to converge properly and failed after three grasps. 
Finally, when not using surgical tool tracking, no attempt could be made successful because the grasper misses the tissue.
All three of these failure cases are shown in Fig. \ref{fig:failure_results}.

    

\begin{table}[tp]
    \centering
    \caption{\\Reprojection Error for Tracking Accuracy}
    \setlength\tabcolsep{1.0em}    
    \begin{tabular}{c|c|c|c}
        Num. of Particle & Mean IoU & Perc. above 80\% & Fps \\
        & & & \\[-1em] \hline 
        & & & \\[-1em]
        100 & 80.8\%& 68\%& 30\\
        & & & \\[-1em] \hline
        & & & \\[-1em]
        500 & 82.4\% & 71\% & 30 \\
        & & & \\[-1em] \hline
        & & & \\[-1em]
        1000 & 81.7\% & 73\% & 26 \\
        & & & \\[-1em] \hline
        & & & \\[-1em]
        5000 & 82.8\% & 77\% & 8
    \end{tabular}   
    \vspace{-0.14in}
    \label{table:results_quantitative}
\end{table}

\begin{figure}[!t]
\centering
\includegraphics[width=0.9\linewidth]{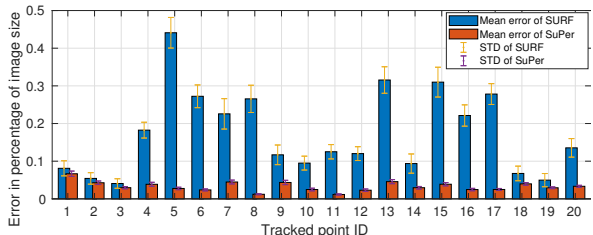}
\vspace{-0.08in}
\caption{The reprojection error comparison of 20 labeled points in our dataset between our SuPer and native SURF keypoint tracking. }
\label{fig:quantitative_experiments}
\vspace{-0.16in}
\end{figure}

A comparison between the reprojected rendering from the surgical tool and the manual segmentation results are shown in Table.~\ref{table:results_quantitative}. We can see that more particles generally gives better performance. However, for efficiency, we set the number as 500 to keep the method in real-time.
In Fig.~\ref{fig:quantitative_experiments}, we can see that our SuPer is much more stable compared to SURF feature matching since our method tries to reconstruct the dynamic scene entirely while SURF only finds the local minimal matching position. 
Also, our method is much more accurate as our error is smaller than SURF even with tracked point No.2 and No.3 which are the best performance of SURF as shown in Fig.~\ref{fig:quantitative_experiments}.



\section{Discussion and Conclusion}

The ability to continuously and accurately track the tissue during manipulation enables control algorithms to be successful in the unstructured environment. 
Currently, we believe that the first limiting factor of our system is the noise from the depth map reconstructed by the stereo-endoscopic camera as shown in Fig.~\ref{fig:depthMap}. 
An experimental second limitation is the features used to update the surgical tool tracker. The markers were manually painted and are inaccurate in terms of position.
We believe this is the main cause of the inconsistency in the surgical tool tracking, and other methods such as as~\cite{Mathisetal2018deeplabcut} would be viable to use in place of the color tracking.
Improving these components would be simple as other strategies for more recent and effective depth reconstruction and instrument feature tracking could be substituted at no additional effort.
Furthermore, the certainty of the perception can be used for optimal control algorithms, endoscopic camera control to maximize certainty, and other advanced control techniques.
Handling blood and topological changes, such as cutting, are the next big challenges to overcome to make our proposed framework even more suitable for real clinical scenarios.


In conclusion, we proposed a surgical perception framework, SuPer, to localize the surgical tool and track the deformable tissue. 
SuPer was evaluated experimentally on a da Vinci\textregistered{} System to show its ability to track under manipulation tasks where instrument occlusions, significant tissue deformations, and tissue tracking were necessary to be handled.
In addition, a deformable tissue tracking dataset was released for further community research.



\section*{ACKNOWLEDGMENT}
This research is supported by the National Natural Science Foundation of China under Grants~(61831015), the fund from Zhejiang University Academic Award for Outstanding Doctoral Candidates, the UCSD Galvanizing Engineering in Medicine~(GEM) program, the GPU grant from Nvidia and the National Science Foundation Graduate Research Fellowships awarded to F. Richter.

\bibliographystyle{ieeetr}
\bibliography{references}

\end{document}